\documentclass[runningheads]{llncs}
\usepackage[T1]{fontenc}
\usepackage{graphicx,verbatim}
\usepackage{subcaption}
\usepackage{amsmath}
\usepackage{booktabs} 
\usepackage{array}
\usepackage{bigstrut}
\usepackage{xcolor}
\usepackage{colortbl}
\usepackage{multirow}
\usepackage{adjustbox}
\usepackage[hyperindex, plainpages=false, colorlinks=true, linkcolor=blue, citecolor=blue, urlcolor=black]{hyperref}
\usepackage{pifont}
\makeatletter
\def\thanks#1{\protected@xdef\@thanks{\@thanks
        \protect\footnotetext{#1}}}
\makeatother

\begin{document}

\title{PerioDet: Large-Scale Panoramic Radiograph Benchmark for Clinical-Oriented Apical Periodontitis Detection}
%

\author{Xiaocheng Fang\inst{1,2}$^{\ast}$ \and
Jieyi Cai\inst{1,2}$^{\ast}$ \and
Huanyu Liu\inst{1,2} \and
Chengju Zhou\inst{2} \and \\
Minhua Lu\inst{3} \and
Bingzhi Chen\inst{1}$^{\dag}$\thanks{$^{\ast}$Equal contribution. \\ $^{\dag}$Corresponding author: Bingzhi Chen}}

\institute{Beijing Institute of Technology, Zhuhai \and
South China Normal University \and Shenzhen University \\
\email{fangxiaocheng@m.scnu.edu.cn, chenbingzhi@bit.edu.cn}}

\maketitle
\markboth{Fang, Cai et al.}{PerioDet}
\begin{abstract}
Apical periodontitis is a prevalent oral pathology that presents significant public health challenges. 
Despite advances in automated diagnostic systems across various medical fields, the development of Computer-Aided Diagnosis (CAD) applications for apical periodontitis is still constrained by the lack of a large-scale, high-quality annotated dataset. 
To address this issue, we release a large-scale panoramic radiograph benchmark called "\textbf{PerioXrays}", comprising 3,673 images and 5,662 meticulously annotated instances of apical periodontitis. To the best of our knowledge, this is the first benchmark dataset for automated apical periodontitis diagnosis.  This paper further proposes a clinical-oriented apical periodontitis detection (\textbf{PerioDet}) paradigm,  
which jointly incorporates Background-Denoising Attention (BDA) and IoU-Dynamic Calibration (IDC) mechanisms to address the challenges posed by background noise and small targets in automated detection. Extensive experiments on the PerioXrays dataset demonstrate the superiority of PerioDet in advancing automated apical periodontitis detection. Additionally, a well-designed human-computer collaborative experiment underscores the clinical applicability of our method as an auxiliary diagnostic tool for professional dentists. The project is publicly accessible at \url{https://github.com/XiaochengFang/MICCAI2025_PerioDet}.


\keywords{Computer-Aided Diagnosis \and Panoramic Radiograph \and Apical Periodontitis Diagnosis \and PerioXrays \and PerioDet.}

\end{abstract}

\section{Introduction}
\begin{figure}[t]
    \centering
    \begin{subfigure}{0.45\linewidth}
        \centering
        \includegraphics[width=0.958\columnwidth]{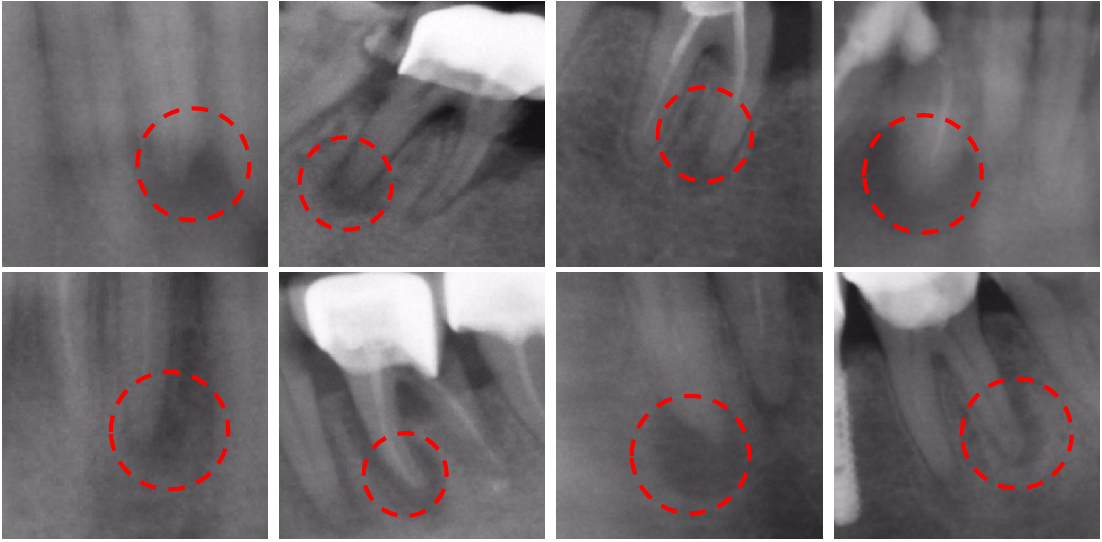}
        \caption{Representative examples}
        \label{Fig_1a}
    \end{subfigure}
    \begin{subfigure}{0.45\linewidth}
        \centering
        \includegraphics[width=0.928\columnwidth]{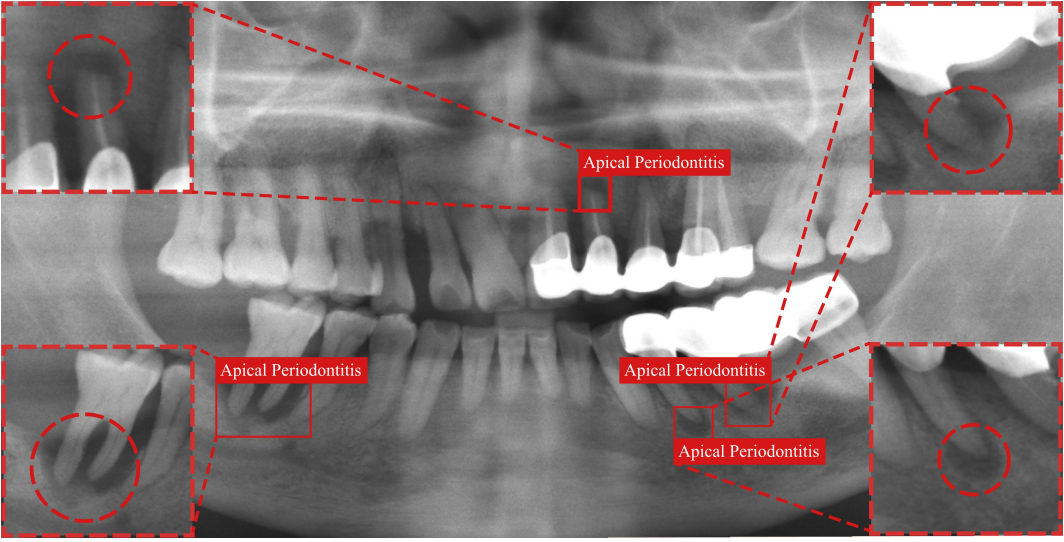}
        \caption{Example with manual annotations}
        \label{Fig_1b}
    \end{subfigure}
    \caption{Apical periodontitis in panoramic radiograph from PerioXrays.}
    \label{Fig_1}
\end{figure}

Apical periodontitis is one of the most prevalent oral pathologies, affecting approximately 52\% of adults worldwide~\cite{tiburcio2021global}. Early and accurate diagnosis is essential for the effective management and planning of endodontic treatments to prevent complications~\cite{estrela2008prevalence,karunakaran2017successful}. Panoramic dental radiography, a standard diagnostic tool, provides a comprehensive view of dental structures, aiding in the detection of apical periodontitis~\cite{choi2022automatic}. However, the accuracy of diagnosis can be affected by the radiologist’s experience and fatigue, particularly during prolonged hours of interpreting panoramic X-rays. This is especially problematic when identifying subtle lesions, as illustrated in Fig.~\ref{Fig_1a}, where apical periodontitis diagnosis remains challenging due to the subtle or ambiguous characteristics of lesions, as well as background noise such as low image quality, poor contrast, and artifacts. Thus, it is crucial to develop an automated system for computer-aided diagnosis of apical periodontitis from panoramic dental X-rays.

Recent advancements in medical image analysis have been greatly influenced by the development of large-scale datasets~\cite{wang2017chestx,rajpurkar2017mura,halabi2019rsna,chen2024cariesxrays,chen2025oralxrays} and deep learning techniques~\cite{lu2021deep,greenwald2022whole,chen2023transattunet}. Comprehensive datasets covering a wide range of medical conditions and imaging modalities have emerged. For example, the ChestXray14 dataset~\cite{wang2017chestx}, containing over 112,000 chest X-ray images, has facilitated deep learning applications for detecting a variety of chest conditions. Similarly, the MURA dataset~\cite{rajpurkar2017mura}, with 40,000 upper limb radiographs, has been pivotal in developing models for diagnosing musculoskeletal diseases, especially bone and joint abnormalities. The RSNA Pediatric Bone Age Challenge dataset~\cite{halabi2019rsna}, consisting of 14,236 pediatric radiographs, aids in training models to assess bone age in children, supporting accurate monitoring of growth and development. These datasets have significantly advanced AI-driven diagnostic systems, demonstrating the potential of deep learning in condition detection, classification, and age assessment. However, progress in AI-based diagnostics for apical periodontitis detection remains limited, mainly due to the absence of large-scale datasets.

Technically, automated detection of apical periodontitis faces two key challenges: \textbf{1) Background noise:} The detection of apical periodontitis is often compromised by background noise arising from factors such as low image quality, poor contrast, and various artifacts in panoramic dental X-rays. These disturbances obscure lesion details, making it difficult to distinguish pathological changes from surrounding anatomical structures, thereby hindering accurate detection.  \textbf{2) Small targets:} Apical periodontitis typically manifests as subtle, localized changes at the root apex with relatively small dimensions in radiographs. This characteristic poses a significant challenge for conventional image analysis methods, as precise identification techniques are required to detect even the slightest structural variations.

To address the above issues, we have developed a large-scale, high-quality annotated benchmark dataset, namely “\textbf{PerioXrays}” for automated apical periodontitis detection. The dataset comprises 3,673 panoramic X-rays with 5,662 meticulously annotated instances of apical periodontitis. Each annotation was reviewed by four experienced professional dentists to ensure accuracy and reliability. Additionally, we propose a clinical-oriented apical periodontitis (\textbf{PerioDet}) paradigm. To mitigate the impact of background noise in panoramic dental X-rays, the Background-Denoising Attention (BDA) module is designed to refine feature representations by modeling channel importance while establishing associations between target and scene features, effectively suppressing irrelevant background interference. To address the challenge of detecting small lesions, we propose the IoU-Dynamic Calibration (IDC) module, which employs an Adaptive IoU Threshold to dynamically adjust positive sample criteria based on lesion size, ensuring a sufficient number of anchors for small lesions. A Dynamic Label Assignment strategy refines anchor-ground truth matching, progressively adjusting sample selection throughout training to enhance localization precision. Our PerioDet paradigm, evaluated on the PerioXrays dataset, achieves state-of-the-art performance and demonstrates clinical potential as an auxiliary diagnostic tool for dentists in human-computer collaborative experiments.

\begin{figure}[t]
    \centering
    \begin{subfigure}{0.32\linewidth}
        \centering
        \includegraphics[width=\linewidth]{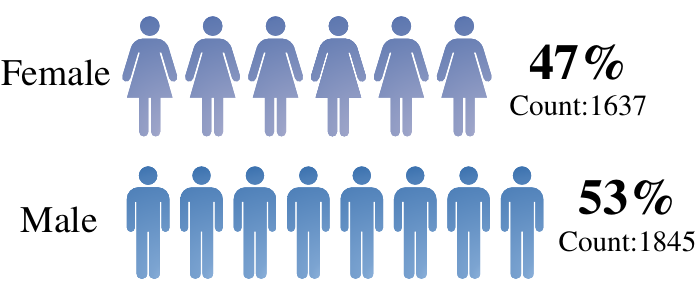}
        \caption{Gender distribution}
        \label{Fig_2a}
    \end{subfigure}
    \begin{subfigure}{0.32\linewidth}
        \centering
        \includegraphics[width=0.75\linewidth]{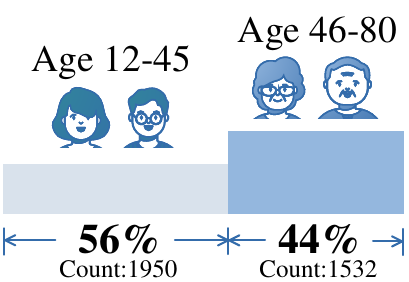}
        \caption{Age distribution}
        \label{Fig_2b}
    \end{subfigure}
    \begin{subfigure}{0.32\linewidth}
        \centering
        \includegraphics[width=\linewidth]{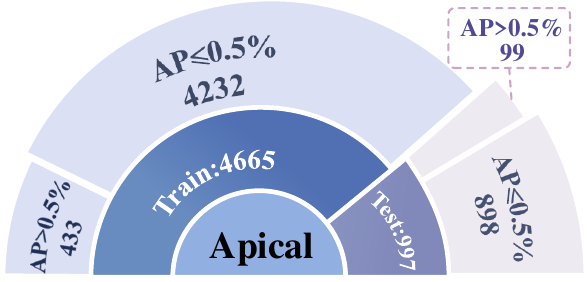}
        \caption{Lesion size distribution}
        \label{Fig_2c}
    \end{subfigure}
    \caption{Comprehensive dataset statistics and distribution about PerioXrays.}
    \label{Fig_2}
\end{figure}

\section{PerioXrays Dataset Creation}
\subsection{Panoramic X-rays Collection}
The PerioXrays dataset is a comprehensive collection of panoramic dental X-rays obtained from multiple hospitals, including dental clinics and orthodontic centers, gathered between 2022 and 2024. As shown in Fig.~\ref{Fig_2a} and Fig.~\ref{Fig_2b}, the dataset covers a wide range of apical periodontitis cases across diverse demographic groups, including various age ranges and genders. All images are standardized to a resolution of $1333 \times 800$ pixels to ensure consistency while minimizing information loss.

\subsection{Apical Periodontitis  Annotations}
In the PerioXrays dataset, a custom annotation tool was used to delineate bounding boxes around instances of apical periodontitis. Each bounding box was precisely adjusted to capture the true extent of the lesion, avoiding both overextension and undersizing. The annotation process followed comprehensive guidelines based on established clinical diagnostic criteria for apical periodontitis~\cite{abbott2004classification,huumonen2002radiological,armitage1995clinical}. Fig.~\ref{Fig_1b} shows post-annotation examples from the dataset, highlighting the identified instances of apical periodontitis. To ensure accuracy, each image underwent a rigorous multi-stage review by four experienced professional dentists.

\subsection{Dataset Distribution Statistics}
The PerioXrays dataset contains 3,673 panoramic dental X-ray images from 3,482 unique patients, with a total of 5,662 annotated instances of apical periodontitis. Fig.~\ref{Fig_2c} shows the distribution of object counts and the area ratio (AR) of labeled instances across the training set and test set. The AR metric represents the proportion of an image occupied by apical periodontitis lesions, indicating their size and spatial coverage. Detecting small lesions is particularly challenging due to their limited visibility and spatial presence. The high prevalence of small objects (AR$\leq$0.5\%) underscores the need for advanced detection algorithms to accurately localize apical periodontitis.

\begin{figure*}[t]
\centering
\includegraphics[width=0.99\linewidth]{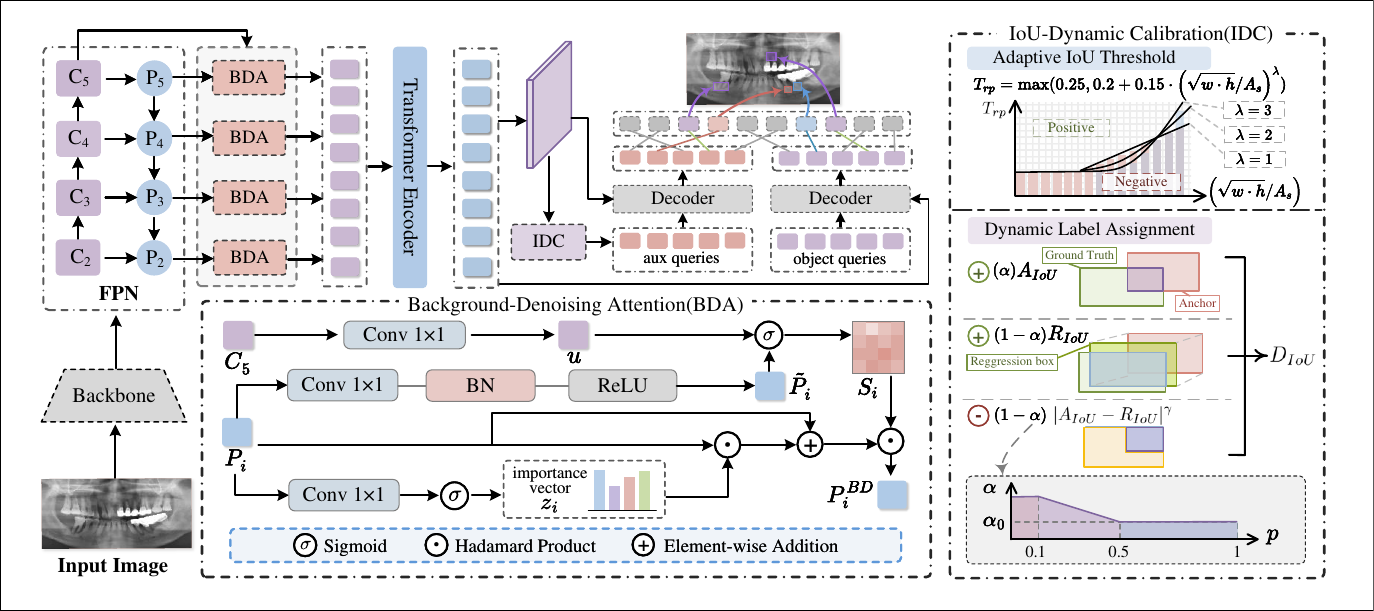}
\caption{Illustration of our proposed PerioDet paradigm, which incorporates BDA and IDC mechanisms for automated apical periodontitis detection accurately.}
\label{Fig_3}
\end{figure*}

\section{Methodology}
\subsection{Background-Denoising Attention}
To mitigate the impact of background noise in panoramic dental X-rays, we propose the Background-Denoising Attention (BDA) module, which enhances feature representations by modeling channel importance while simultaneously capturing associations between target and scene features, effectively suppressing irrelevant background noise.

As shown in Fig.~\ref{Fig_3}, $P_i$ represents the features in the top-down path in Feature Pyramid Network(FPN)~\cite{lin2017feature}. The BDA module generates a new feature map, \( P^{BD}_i \), by weighting \( P_i \) using the importance vector \( z_i \) and the similarity map \( S_i \). The importance vector \( z_i \) is computed as follows,
\begin{equation}
\label{Eq_1}
z_i = \text{Sigmoid}\left(\text{Conv}_{1 \times 1}(P_i)\right),
\end{equation}
which is designed to prioritize feature channels that capture rich spatial details while suppressing background noise.
 
The similarity map \( S_i \) captures the relationship between the target and scene features, effectively suppressing irrelevant background interference. To compute \( S_i \), two projection functions are employed to learn the target and scene features. 

The projection of the target feature involves applying a series of operations to \( P_i \), including a \( 1 \times 1 \) convolution for dimensionality reduction, followed by Batch Normalization and ReLU activation to enhance non-linearity and stabilize training. This process is formulated as follows,
\begin{equation}
\label{Eq_2} 
\tilde{P}_i = \text{ReLU}(\text{BN}(\text{Conv}_{1 \times 1}(P_i))).
\end{equation}

Additionally, a one-dimensional scene embedding vector \( u \) is computed to capture the contextual information of the scene. This embedding vector is generated by applying a \(1\times1 \) convolution operation to the feature map \( C_5 \), effectively projecting the scene features into a compact representation while preserving essential contextual details. The computation of \( u \) is expressed as follows,
\begin{equation}
\label{Eq_3} 
u = \text{Conv}_{1 \times 1}(C_5).
\end{equation}

With the refined feature map \( \tilde{P}_i \) and the scene embedding vector \( u \), the similarity map \( S_i \) is calculated as follows,
\begin{equation}
\label{Eq_4} 
S_i = \text{Sigmoid}(\tilde{P}_i \cdot u ).
\end{equation}

Based on the above analyses, the final output is computed as follows,
\begin{equation}
\label{Eq_5} 
P^{BD}_i = (1 + z_i) \cdot P_i \odot S_i.
\end{equation}

\subsection{IoU-Dynamic Calibration}
To address the problem posed by small targets in apical periodontitis detection,  the IoU-Dynamic Calibration(IDC) module mainly benefits from the Adaptive IoU Threshold and  Dynamic Label Assignment mechanisms.

\textbf{Adaptive IoU Threshold.} 
Traditional IoU thresholds struggle to capture small lesions due to their sensitivity to positional deviations\cite{dong2023control}. To address this limitation, we propose an Adaptive IoU Threshold that dynamically adjusts the positive sample thresholds based on the object's area, ensuring sufficient anchors for small lesions. The Adaptive IoU Threshold is defined as follows,
\begin{equation}
\label{Eq_6}
T_{rp} = \max(0.25, 0.2 + 0.15 \cdot \left(\sqrt{w \cdot h}/ {A_s}\right)^\lambda),
\end{equation}
where $w$ and $h$ denote the width and height of the object, and $A_s$ represents the minimal area definition which is adjusted for different datasets. $T_{rp}$ is the reference threshold for positive samples, while $\lambda$ controls the growth rate of $T_{rp}$.

\textbf{Dynamic Label Assignment.} To enhance label assignment for small lesions, we propose a Dynamic Label Assignment that employs Dynamic IoU(DIoU) to adaptively adjust the sample selection during training, refining the matching process between anchors and ground truth (GT). DIoU is defined as follows,
\begin{equation}
\label{Eq_7}
D_{IoU} = \alpha \cdot A_{IoU} + (1 - \alpha) \cdot R_{IoU} - (1 - \alpha) \cdot |A_{IoU} - R_{IoU}|^\gamma,
\end{equation}
where \( A_{IoU} \) represents the IoU of anchor and GT, \( R_{IoU} \) represents the IoU of regression box and GT, and $\alpha$ and $\gamma$ are hyperparameters used for weighting.

Our Dynamic Label Assignment strategy follows the following procedure. First, we calculate $T_{rp}$ based on Eq.~\ref{Eq_6}. Subsequently, the DIoU of each anchor is computed during label assignment, and anchors are designated as positive samples when DIoU is greater than $T_{rp}$. To ensure the stability of training, the influence of $R_{IoU}$ on sample selection is gradually increased with training iterations. The adjustment of \(\alpha\) at each stage follows the criteria:
\begin{equation}
\label{Eq_8}
\alpha(p, \alpha_0) =
\begin{cases} 
1, & 0 \leq p < 0.1 \\ 
\left( \frac{\alpha_0 - 1}{0.5 - 0.1} \right) (p - 0.1) + 1, & 0.1 \leq p < 0.5 \\ 
\alpha_0, & p \geq 0.5 
\end{cases}
\end{equation}
where p = $\frac{\text{epoch}}{\text{epochs}}$, with \(\text{epoch}\) representing the current training epoch and \(\text{epochs}\) the total number of epochs. The hyperparameter $\alpha_0$ controls $\alpha$.

\section{Experiments}
\subsection{Experiment Setup}
To evaluate our method, we compare various object detection baselines on the PerioXrays dataset, including \textbf{CNN-based} and \textbf{Transformer-based} approaches. The dataset was randomly split into a training set of 3,000 images and a test set of 673 images, with the division made at the patient level to ensure that all images from a single patient were assigned to the same set. We use Average Precision (AP) \cite{lin2014microsoft} as the primary metric, along with $\text{AP}_{50}$ and $\text{AP}_{75}$ at IoU thresholds of 0.5 and 0.75. Objects are categorized into small (S), medium (M), and large (L) sizes for detailed performance analysis. During training, we used AdamW as the optimizer with a batch size of 2, a 1$\times$ schedule (12 epochs), an initial learning rate of 2e-4, and a weight decay of 0.0001. All baseline models were trained with a \textbf{ResNet-50 backbone} for fair comparison.

\begin{table*}[t]
\small
\centering
\renewcommand\tabcolsep{1.2pt}
\renewcommand\arraystretch{1.0}
\caption{Comparison with detection approaches on the PerioXrays.}
\begin{tabular}{c|l|c|cccccc}
    \toprule
    \textbf{Models}&
    \textbf{Methods}& \textbf{Ref.} &\textbf{$\text{AP}$} &\textbf{$\text{AP}_{50}$}   & \textbf{$\text{AP}_{75}$}& \textbf{$\text{AP}_{S}$}& \textbf{$\text{AP}_{M}$}&\textbf{ $\text{AP}_{L}$}\bigstrut[b]\\
    \hline
    \multirow{6}{*}{\textbf{CNN}}&
    Faster RCNN~\cite{ren2015faster}& NIPS'15 &46.1 & 76.4 & 50.1 & 23.6 & 44.7 & 53.1\\
    &ATSS~\cite{zhang2020bridging}& CVPR'20 &47.2 & 80.2 & 50.1 & 33.0 & 46.3 & 54.4\\
    &YOLOX~\cite{ge2021yolox}& CVPR'21 &46.7 & 79.5 & 48.4 & 28.3 & 45.2 & 55.1\\
    
    &Sparse RCNN~\cite{sun2021sparse}& CVPR'21 &47.0 & 80.1 & 49.5 & 36.9 & 45.1 & 54.3\\
    
    &RFLA~\cite{xu2022rfla}& ECCV'22 &47.5 &80.3 &50.6 &36.4 &46.1 &55.0 \\
    &CFINet~\cite{yuan2023small}& ICCV'23 &47.7 &80.6 &51.0 &\underline{37.2} &46.4 &55.6\\
    \hline
    \multirow{7}{*}{\textbf{Transformer}}&
    Deformable-DETR~\cite{zhu2020deformable}& ICLR'20 &38.4 & 75.8 & 35.1 & 21.7 & 36.6 & 46.6\\ 
    &Conditional-DETR~\cite{meng2021conditional}& ICCV'21 & 42.6 &  78.4 & 41.0 & 23.3 & 41.1 & 49.7\\ 
    &DAB-DETR~\cite{liu2021dab}& ICLR'22 & 44.0 & 78.5 & 45.0 & 33.4 & 43.1 & 50.0\\ 
    &DINO~\cite{zhang2022dino}& ICLR'23 & 48.9 & 81.2 & 51.3 & 20.4 & 47.6 & 56.9\\ 
    &Co-DINO~\cite{zong2023detrs}& ICCV'23 &49.9 & 81.5 & 53.4 &36.8 & \underline{48.5} & 57.7\\
    &Salience-DETR~\cite{hou2024salience} & CVPR'24 & \underline{50.3} & \underline{81.7} & \underline{53.7} & 37.0 & 48.2 &\underline{57.8} \\ & \cellcolor{gray!20}\textbf{Ours}& \cellcolor{gray!20}-&\cellcolor{gray!20}\textbf{53.5} & \cellcolor{gray!20}\textbf{84.2} & \cellcolor{gray!20}\textbf{55.6} & \cellcolor{gray!20}\textbf{42.3} & \cellcolor{gray!20}\textbf{50.9} & \cellcolor{gray!20}\textbf{58.7}\\
    \bottomrule
\end{tabular}
\label{Table1}
\end{table*}

\subsection{Comparisons with State-of-The-Arts} 
We evaluate the performance of our method on the PerioXrays dataset, where it surpasses all baseline approaches. As shown in Table~\ref{Table1}, PerioDet achieves a 5.8\% improvement in AP compared to leading CNN-based models such as CFINet (53.5\% vs. 47.7\%) and a 3.2\% improvement over state-of-the-art Transformer-based models like Salience-DETR (53.5\% vs. 50.3\%). Notably, PerioDet excels in detecting small lesions, with a 5.1\% improvement in AP$_{S}$.

\subsection{Ablation Studies and Parameter Analysis} 
\begin{figure}[t]
    \centering
    \begin{subfigure}{0.30\linewidth}
        \centering
        \includegraphics[width=\linewidth]{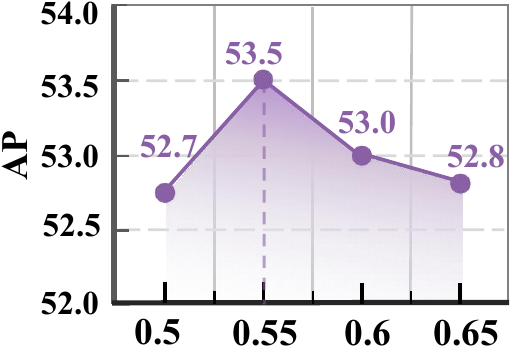}
        \caption{Parameter $\lambda$}
        \label{Fig_4a}
    \end{subfigure}
    \begin{subfigure}{0.30\linewidth}
        \centering
        \includegraphics[width=\linewidth]{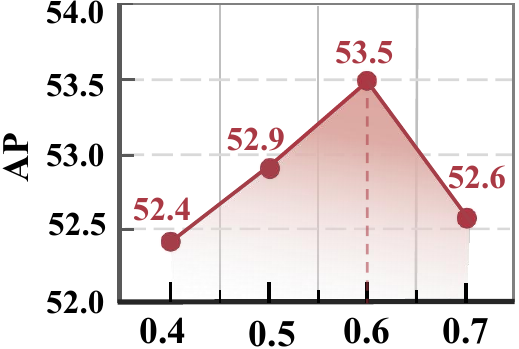}
        \caption{Parameter $\alpha_0$}
        \label{Fig_4b}
    \end{subfigure}
    \begin{subfigure}{0.30\linewidth}
        \centering
        \includegraphics[width=\linewidth]{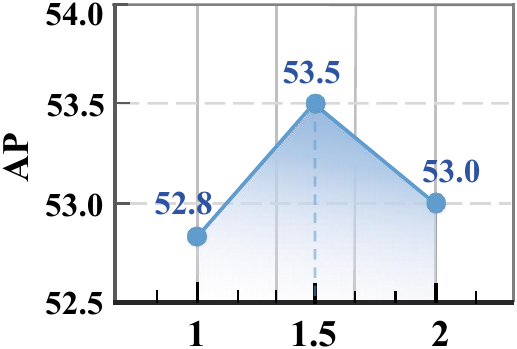}
        \caption{Parameter $\gamma$}
        \label{Fig_4c}
    \end{subfigure}
    \caption{Parameter analysis (\%) on the PerioXrays dataset.}
    \label{Fig_4}
\end{figure}

In our ablation studies, we systematically assess the contribution of each component in the PerioDet method using the PerioXrays dataset to validate its effectiveness. As shown in Table~\ref{Table4}, the results demonstrate that all modules work synergistically, reinforcing each other and confirming the overall effectiveness of the method. Additionally, we analyze the impact of three critical hyperparameters, $\lambda$, $\alpha_0$, and $\gamma$, as defined in Eq.~\ref{Eq_6}, Eq.~\ref{Eq_7} and Eq.~\ref{Eq_8}. Fig.~\ref{Fig_4} shows that our PerioDet achieves optimal performance with $\lambda=0.55$, $\alpha_0=0.6$, and $\gamma=1.5$.

\begin{table}[t]
\small
\centering
\begin{minipage}{0.43\linewidth} 
    \centering
    \renewcommand\tabcolsep{0.2pt}
\renewcommand\arraystretch{0.8}
    \caption{Ablation studies.}
\begin{tabular}{cc|ccc|ccc}
         \toprule
          \textbf{BDA}&\textbf{IDC}&\textbf{$\text{AP}$} &\textbf{$\text{AP}_{50}$} & \textbf{$\text{AP}_{75}$}& \textbf{$\text{AP}_{S}$}& \textbf{$\text{AP}_{M}$}& \textbf{$\text{AP}_{L}$}\\
          \midrule
          \ding{55}&\ding{55} &49.9 &81.5&53.4&36.8&48.5&57.7   \\
          \ding{51} &\ding{55}&51.7&83.1&54.7&40.6&49.8&58.0 \\
          \ding{55}&\ding{51}&52.3&83.5&55.2&41.1&50.3&58.3\\ 
          \ding{51}&\ding{51}&53.5&84.2&55.6&42.3&50.9&58.7\\
          \bottomrule
     \end{tabular}
    \label{Table4}
\end{minipage}
\hfill
\begin{minipage}{0.49\linewidth} 
    \centering
    \renewcommand\tabcolsep{0.1pt}
\renewcommand\arraystretch{1.015}
    \caption{Diagnostic comparison.}
    \begin{tabular}{c|c|c}
        \toprule
        \textbf{Methods} & 
        \textbf{w/o PerioDet}&
        \textbf{w PerioDet} \\
        \midrule
        \textbf{Precision} & 73.1\% & 92.5\%(\textcolor{red}{19.4\% $\uparrow$})\\
        \textbf{Recall} & 74.3\% & 96.1\%(\textcolor{red}{21.8\% $\uparrow$})\\
        \textbf{Efficiency} & $\approx$28 s/img & $\approx$13 s/img(\textcolor{red}{15s$\downarrow$})\\
        \bottomrule
    \end{tabular}
    \label{Table5}
\end{minipage}
\end{table}

\begin{figure*}[t]
\centerline{\includegraphics[width=\textwidth]{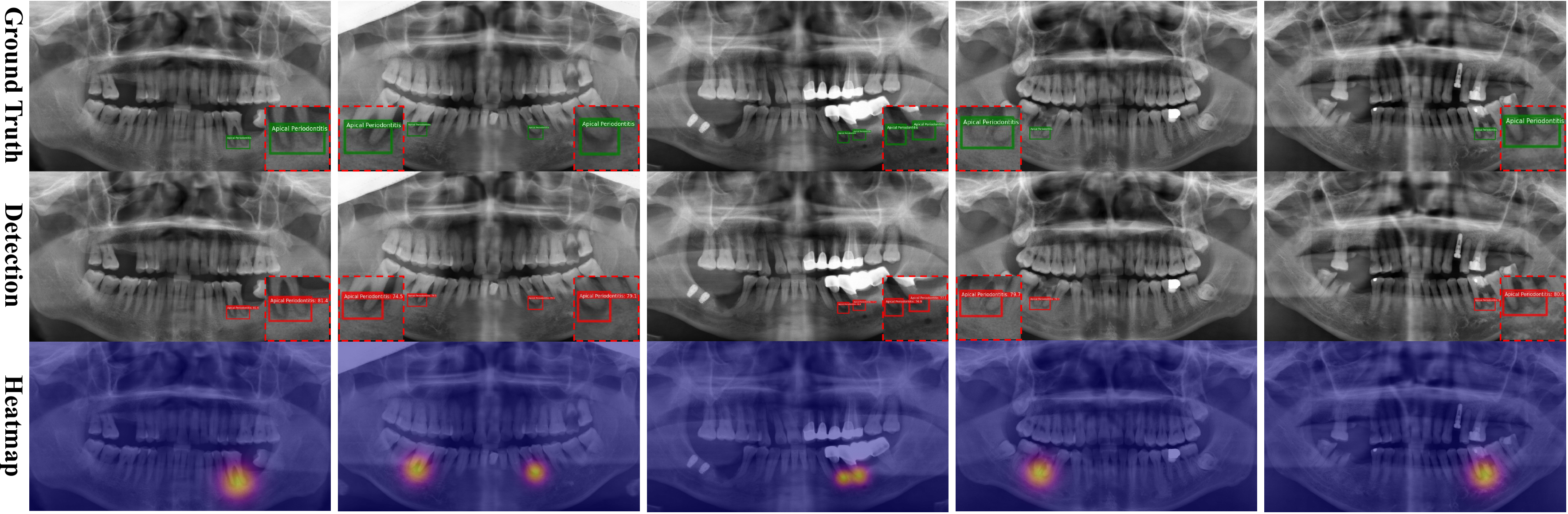}}
\caption{Visualization of detections and heatmaps on the PerioXrays dataset.}
\label{Fig_6}
\end{figure*}

\begin{figure*}[!t]
\centerline{\includegraphics[width=\textwidth]{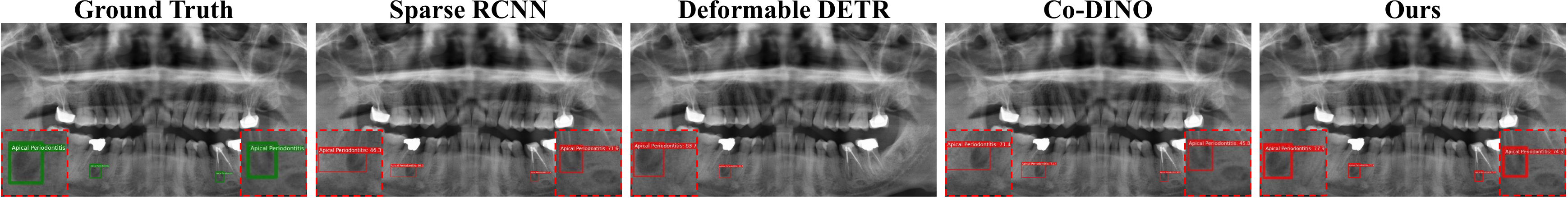}}
\caption{Comparison of predicted bounding boxes across various models.}
\label{Fig_7}
\end{figure*}

\subsection{Human-Computer Collaborative Experiment} 
We designed a novel human-computer collaborative experiment to evaluate the clinical applicability of our method as an auxiliary diagnostic tool. Six professional dentists participated, including two junior, two mediate, and two senior professionals. None had prior exposure to the data, ensuring an unbiased evaluation. We randomly selected 100 panoramic X-ray images from PerioXrays as the test set, and participants made two consecutive diagnostic judgments for each image. As shown in Table \ref{Table5}, these results highlight the clinical applicability of PerioDet, demonstrating its effectiveness in real-world medical diagnostics.

\subsection{Visualization Results} 
We visualize detections and heatmaps. As shown in Fig.~\ref{Fig_6}, PerioDet accurately identifies apical periodontitis and effectively localizes the lesions. The attention distribution further emphasizes PerioDet's ability to capture relevant features, highlighting its enhanced sensitivity to regions of interest. Additionally, Fig.~\ref{Fig_7} demonstrates that PerioDet achieves higher confidence scores and successfully detects apical periodontitis that other models miss.

\section{Conclusion}
In this paper, we release a large-scale panoramic radiograph benchmark called "PerioXrays". Moreover, we propose a clinical-oriented apical periodontitis detection (PerioDet) paradigm which jointly incorporates Background-Denoising Attention (BDA) and IoU-Dynamic Calibration (IDC) mechanisms to address the challenges posed by background noise and small targets. Extensive experiments on the PerioXrays dataset, along with a well-designed human-computer collaborative experiment, establish a strong foundation for automated apical periodontitis detection in dental healthcare.

\subsubsection{\ackname}  
This work was supported in part by the Guangdong Basic and Applied Basic Research Foundation (No. 2025A1515010225), and in part by the National Natural Science Foundation of China (No. 62302172).

\subsubsection{\discintname} The authors have no competing interests to declare that are relevant to the content of this article.

\bibliographystyle{splncs04}
\bibliography{MICCAI2025references.bib}
\end{document}